\documentclass{article} % For LaTeX2e
\usepackage{iclr2026_preprint,times}

\usepackage{amsmath,amsfonts,bm}

\def\eqref#1{equation~\ref{#1}}
\def\1{\bm{1}}

\DeclareMathAlphabet{\mathsfit}{\encodingdefault}{\sfdefault}{m}{sl}
\SetMathAlphabet{\mathsfit}{bold}{\encodingdefault}{\sfdefault}{bx}{n}

\PassOptionsToPackage{numbers,compress}{natbib}
\usepackage{graphicx}
\usepackage{amsthm}
\theoremstyle{plain}

\theoremstyle{definition}

\usepackage{wrapfig}
\usepackage{algorithm}
\usepackage{algorithmic}
\usepackage{tabularray}
\usepackage{amssymb}
\usepackage{subcaption}
\usepackage{multirow}
\usepackage{arydshln}
\usepackage{tabularx}
\usepackage[utf8]{inputenc} % allow utf-8 input
\usepackage[T1]{fontenc}    % use 8-bit T1 fonts
\usepackage{hyperref}       % hyperlinks
\usepackage{url}            % simple URL typesetting
\usepackage{booktabs}       % professional-quality tables
\usepackage{amsfonts}       % blackboard math symbols
\usepackage{nicefrac}       % compact symbols for 1/2, etc.
\usepackage{microtype}      % microtypography
\usepackage{xcolor}         % colors
\usepackage{arydshln}

\title{Role-Conditioned Sub-Token Routing for Efficient Vision-Language-Action Policies}

\author{Wei Jiang \\
Futurewei Technologies\\
San Jose, CA 95131, USA \\
\texttt{\{wjiang@futurewei.com} \\
\And
Wei Wang \\
Futurewei Technologies\\
San Jose, CA 95131, USA \\
\texttt{rickweiwang@futurewei.com} \\
}

\iclrfinalcopy % Uncomment for camera-ready version, but NOT for submission.
\begin{document}

\maketitle

\begin{abstract}
Vision-Language-Action (VLA) models process long multimodal token sequences,
making inference expensive in both memory and computation. Existing efficiency
methods mainly reduce visual tokens, but aggressive token pruning becomes
fragile because removing a token discards its entire representation. Sub-token
compression provides a complementary alternative by retaining more tokens
while reducing their value width. However, directly applying
sub-token compression to VLA policies is less effective because information
important for perception, language understanding, and control is distributed
differently across the multimodal representation.

We introduce Role-Conditioned Sub-Token Routing (RoleSub), which
learns how to compress the value representations of retained tokens. After
visual token reduction, RoleSub partitions each retained value representation
into groups in an orthogonal space and uses a lightweight router to determine
which groups should be preserved. The routing decision is conditioned on the
token representation, a learned latent role representation, and language
context. The same mechanism can also be applied to language values, allowing
visual and language representations to be compressed without removing
additional tokens.

We evaluate RoleSub on OpenVLA-OFT-7B across the four LIBERO suites. At
matched visual-KV budgets, RoleSub outperforms a trained token-only control in
33 of 36 settings, with the largest gains under aggressive compression.
Combining visual and language compression reduces total KV to 9.2--11.3\% of
the original while retaining strong control performance on most tasks. These
results show that reducing the representation within retained tokens provides
an effective complement to token pruning for aggressive VLA compression.
\end{abstract}

\section{Introduction}
\label{sec:intro}

Vision-Language-Action (VLA) models have emerged as a promising approach to general-purpose robot control by transferring the perception and reasoning capabilities of large vision-language models (VLMs) to action prediction. Models such as OpenVLA~\cite{kim2025openvla} and OpenVLA-OFT~\cite{kim2025oft} jointly process visual observations, natural-language instructions, and robot states to generate manipulation actions, achieving strong performance across a broad range of tasks. This capability, however, comes with the computational cost of a large multimodal transformer. At every control step, the policy processes a long prefix dominated by visual tokens, together with language, proprioceptive, and action-related states. Reducing the cost of this multimodal context is therefore important for practical VLA systems.

Existing VLA acceleration methods mainly operate at the token level. VLA-ADP~\cite{pei2025vlaadp} and VLA-Pruner~\cite{liu2025vlapruner}, for example, identify visual tokens that are less relevant to the current task and remove them from subsequent computation. Similar strategies have been extensively studied for VLMs, including FastV~\cite{chen2024fastv}, VisionZip~\cite{yang2025visionzip}, SparseVLM~\cite{zhang2025sparsevlm}, and FitPrune~\cite{ye2025fitprune}. These methods exploit the considerable redundancy among visual tokens and are effective at moderate keep ratios. Their compression unit is an entire token. Once a token is removed, all information carried by that token is lost.

This granularity becomes increasingly restrictive as the budget decreases. A visual token that appears only moderately important may still contain information needed for object localization, geometric relations, or subsequent manipulation. Removing the token eliminates all of these signals simultaneously. The problem is amplified in closed-loop control: an incorrect action changes the physical state and therefore changes the observations presented to the policy at later steps. Compression errors that may cause only a local prediction error in a VLM can therefore propagate through an entire VLA trajectory.

Sub-token routing provides a complementary way to reduce transformer KV states~\cite{jiang2026subtoken}. Rather than further reducing the number of retained tokens, it compresses the value representation within each retained token by partitioning the value vector into groups and selecting only a subset of them. Query and key representations remain unchanged. Token-level reduction and sub-token routing therefore act along two different dimensions: one controls how many token states remain, while the other controls how much value information is retained for each state. In LLMs and VLMs, combining these two dimensions yields better accuracy--KV trade-offs than relying on token removal alone, particularly at aggressive compression levels~\cite{jiang2026subtoken}.

Directly transferring the sub-token routing to VLA policies, however, leads to a very different result. In an initial experiment, uniform \(S\!=\!4\), \(K\!=\!1\) sub-token routing applied after visual-token pruning achieves only 43.8\% success on LIBERO-Spatial. It indicates that aggressive VLA compression is not determined only by how much information is retained, but also by how that capacity is distributed within the multimodal representation.

This difference is natural for a VLA policy. Visual, language, proprioceptive, and action-related states participate differently in action generation, and even tokens of the same modality may contribute different information depending on the task and control stage. At the same time, these functions are not expected to occupy known or explicitly separable coordinates of the transformer representation. It is  difficult to predetermine in advance which dimensions correspond to semantics, geometry, temporal context, or control. A fixed decomposition can easily remove dimensions that become important after interaction with the rest of the network. What is needed instead is a representation in which the routing structure itself can be learned from the control objective.

This paper presents \textbf{Role-Conditioned Sub-Token Routing (RoleSub)}, which extends sub-token compression with a learned latent decomposition for VLA policies. For each retained token, the value representation is first transformed into a fixed orthogonal space and partitioned into groups. A lightweight role estimator produces a soft latent representation that participates in determining which groups are retained. Importantly, these latent components are not assumed to correspond to predefined or semantically identifiable roles. Their organization is learned jointly with the policy, allowing the model to discover routing structure that is useful for action prediction. Group selection also depends on the token representation and the language context, while separate budgets are assigned to different token types. The orthogonal transformation provides a lossless change of basis, so the representation is altered only through group selection. This gives the router a structured space in which to learn how limited value capacity should be allocated.

On OpenVLA-OFT-7B across the four LIBERO suites~\cite{liu2023libero},
standard training-free token pruning largely collapses at the aggressive
budgets considered in this work. Since RoleSub is trained jointly with the
compressed policy, comparing only against training-free pruning would not
separate the benefit of sub-token routing from the benefit of adaptation.
We therefore construct a matched-budget token-only control using the same
training procedure as RoleSub, but without sub-token routing. RoleSub outperforms this stronger control in 33 of 36 visual-KV configurations, with the largest gains appearing under aggressive compression and on long-horizon tasks.

The experiments also reveal a strong asymmetry between token-level and sub-token compression of language: removing language tokens rapidly degrades control performance, whereas retaining all language tokens and compressing their value representations to one of sixteen groups produces no measurable loss when applied independently. This suggests that token retention and value-width retention play different roles in VLA policies.

Combining visual and language routing reduces the complete multimodal-prefix KV to 9.2--11.3\% of its original size while retaining strong performance on most LIBERO suites. Long-horizon control remains more sensitive to aggressive language compression, indicating that the appropriate value budget depends on both token type and task demands.

The main contributions of this work are:
\begin{itemize}

\item We introduce RoleSub, which combines an orthogonal value-space decomposition, a learned latent role representation, and token-type-dependent budgets to adapt sub-token routing to VLA control. The latent decomposition is learned jointly from the control objective and does not require predefined or explicitly separable semantic roles.

\item Under matched visual-KV budgets, RoleSub outperforms a trained
token-only control in 33 of 36 LIBERO settings, with the largest gains
appearing under aggressive compression and on long-horizon tasks.

\item We show that token retention and value-width retention behave differently in VLA policies. In particular, language tokens are highly sensitive to token removal, while their value representations can be compressed by up to $16\times$ without measurable loss when compressed independently. Combining visual and language routing further reduces the full multimodal-prefix KV to 9.2--11.3\%.

\end{itemize}

\section{Related Work}
\label{sec:related}

\paragraph{Vision-language-action models.}
Vision-Language-Action models extend pretrained vision-language representations to robot control by conditioning action prediction on visual observations, language instructions, and robot states. OpenVLA~\cite{kim2025openvla} is a 7B-parameter open VLA built on a Llama-2 language model with fused DINOv2 and SigLIP visual features and trained on large-scale robot demonstrations. OpenVLA-OFT~\cite{kim2025oft} improves task adaptation and inference through parallel action decoding, action chunking, continuous action prediction, and $\ell_1$ regression. OpenVLA-OFT serves as the main policy in this work because it provides a strong LIBERO baseline while retaining the transformer structure needed to study multimodal KV compression.

\paragraph{Visual token reduction for VLMs and VLAs.}
A common approach to reducing multimodal transformer cost is to shorten the visual-token sequence. In VLMs, FastV~\cite{chen2024fastv} prunes visual tokens according to attention patterns observed in early layers, while VisionZip~\cite{yang2025visionzip} selects informative visual tokens and removes redundant ones. SparseVLM~\cite{zhang2025sparsevlm} and FitPrune~\cite{ye2025fitprune} similarly exploit text relevance or attention structure to reduce visual-token redundancy. More recent methods adapt token reduction to VLA policies. VLA-ADP~\cite{pei2025vlaadp} combines text-guided token importance with action-aware pruning, while VLA-Pruner~\cite{liu2025vlapruner} incorporates both semantic relevance and action-related information into visual-token selection. These methods operate primarily at token granularity: efficiency is obtained by deciding which visual tokens should remain. RoleSub instead operates after token selection and reduces the value representation within retained tokens, providing a complementary compression axis when further token removal becomes costly.

\paragraph{KV-cache and sub-token compression.}
KV-cache compression has been studied extensively for long-context language models. H$_2$O~\cite{zhang2023h2o} retains recent and heavy-hitter tokens, StreamingLLM~\cite{xiao2024streamingllm} combines attention sinks with a sliding context, and Quest~\cite{tang2024quest} performs query-aware selection of relevant cached states. Other methods reduce the stored representation itself. MiniCache~\cite{liu2024minicache}, for example, exploits similarity between KV states across neighboring layers. Together, these approaches show that useful KV information is distributed nonuniformly across tokens, queries, and model components.

Sub-token routing~\cite{jiang2026subtoken} introduces a complementary form of compression by reducing the value width within retained tokens. Rather than removing additional tokens, it partitions each retained value vector into groups and selects only a subset while leaving the query and key paths unchanged. Experiments on LLMs and VLMs show that this mechanism can complement token-level reduction, particularly at small KV budgets. The VLA setting considered here introduces a different challenge: directly applying uniform sub-token routing can severely degrade closed-loop control. RoleSub addresses this setting by introducing an orthogonal routing space, a learned latent role representation, and token-type-dependent value budgets, without assuming that the learned latent components correspond to predefined semantic roles.

\paragraph{Parameter-efficient adaptation.}
Parameter-efficient fine-tuning provides a practical way to adapt large pretrained models without updating the full backbone. LoRA~\cite{hu2022lora} represents weight updates through low-rank factors and has been widely used for efficient adaptation of large transformers. RoleSub trains the routing components together with low-rank adaptation while keeping the pretrained backbone frozen. The same adaptation procedure is also used for the matched token-only control, allowing the effect of sub-token routing to be separated from the benefit of adapting the policy to compression.

\section{Method}
\label{sec:method}

RoleSub combines token-level reduction with learned sub-token routing for
multimodal VLA representations. Token-level reduction controls which token
states remain in the sequence, while sub-token routing controls how much value
information is retained within each surviving state. These two operations act
along different dimensions and can be configured independently for different
token types. For each token selected for sub-token routing, RoleSub transforms
its value representation into an orthogonal routing space and selects a subset
of value groups according to the token representation, a learned latent role
representation, and the language context. 

Consider a transformer-based VLA policy with $L$ attention layers. At each
control step, the policy receives visual observations, a language instruction,
proprioceptive state, and action-related context. We write the multimodal
sequence as
\begin{equation}
X =
\left[
X^{\mathrm{vis}},
X^{\mathrm{lang}},
X^{\mathrm{prop}},
X^{\mathrm{act}}
\right].\nonumber
\end{equation}
Let $\mathbf{h}_i^{(\ell)} \in \mathbb{R}^{d}$ denote the hidden state of
token $i$ at layer $\ell$, with query, key, and value representations
\begin{equation}
\mathbf{q}_i^{(\ell)}
=
W_q^{(\ell)}\mathbf{h}_i^{(\ell)},\qquad
\mathbf{k}_i^{(\ell)}
=
W_k^{(\ell)}\mathbf{h}_i^{(\ell)},\qquad
\mathbf{v}_i^{(\ell)}
=
W_v^{(\ell)}\mathbf{h}_i^{(\ell)}.\nonumber
\end{equation}
Following sub-token routing~\cite{jiang2026subtoken}, RoleSub keeps the query
and key representations unchanged and applies sub-token compression to the value path. That is, token-level reduction changes the number of token states presented
to subsequent layers, and value routing changes the amount of value information retained within those states.

\subsection{Token-Level Reduction}
\label{sec:tokenprune}

Let
$
m \in
\{
\mathrm{vis},
\mathrm{lang},
\mathrm{prop},
\mathrm{act}
\}
$
denote a token type, and let $\mathcal{I}_m$ denote the set of tokens of
that type. For a token type on which token-level reduction is applied, a
selection function assigns each token an importance score
\begin{equation}
u_i
=
\psi_m
\left(
\mathbf{h}_i,
X
\right),
\qquad
i \in \mathcal{I}_m .\nonumber
\end{equation}

Given a token-retention ratio $r_m$, the selector retains the
$\left\lceil r_m |\mathcal{I}_m| \right\rceil$ highest-scoring tokens,
forming the retained set $\mathcal{I}_m^{\mathrm{keep}}$, where
$\left\lceil\cdot\right\rceil$ denotes the ceiling function. When token-level reduction is not applied to type $m$, $r_m=1$ and all tokens in $\mathcal{I}_m$ are retrained. RoleSub does not require a particular
token-selection rule and can therefore be combined with different token-level reduction methods.

Token retention and sub-token routing are controlled independently. For example, A token type may retain all of its tokens while still compressing their value
representations. This allows RoleSub to reduce token count and value width as two distinct compression dimensions.

\subsection{Orthogonal Value-Space Decomposition}
\label{sec:orthogonal}

For each retained token selected for sub-token routing, RoleSub first maps its
value representation into an orthogonal routing space. Let
$\mathbf{v}_i^{(\ell)} \in \mathbb{R}^{d_v}$ denote the value vector of token
$i$ at layer $\ell$, where $d_v$ is the dimensionality $\mathbf{v}_i^{(\ell)} $. For each layer, we use a fixed orthonormal matrix
\begin{equation}
R^{(\ell)} \in \mathbb{R}^{d_v \times d_v},
\qquad
R^{(\ell)\top}R^{(\ell)} = I,\nonumber
\end{equation}
where $I$ is the $d_v\times d_v$ identity matrix. The value vector is
transformed as
\begin{equation}
\mathbf{z}_i^{(\ell)}
=
R^{(\ell)\top}\mathbf{v}_i^{(\ell)},\nonumber
%\label{eq:orthogonal_projection}
\end{equation}
where $\mathbf{z}_i^{(\ell)} \in \mathbb{R}^{d_v}$ is the value
representation in the orthogonal routing space. We divide $\mathbf{z}_i^{(\ell)}$ into $S$ non-overlapping groups,
\begin{equation}
\mathbf{z}_i^{(\ell)}
=
\left[
\mathbf{z}_{i,1}^{(\ell)}
\mid
\mathbf{z}_{i,2}^{(\ell)}
\mid
\cdots
\mid
\mathbf{z}_{i,S}^{(\ell)}
\right].\nonumber
\end{equation}
Assuming equal-sized groups, each
$\mathbf{z}_{i,s}^{(\ell)} \in \mathbb{R}^{d_s}$ has dimension
$d_s=d_v/S$.

For token $i$, the router retains $K_i$ of the $S$ groups. Let 
$
\mathbf{m}_i^{(\ell)}
=
\left[
m_{i,1}^{(\ell)},\ldots,m_{i,S}^{(\ell)}
\right]
\in \{0,1\}^{S}
$
denote the group-selection mask, where $m_{i,s}^{(\ell)}=1$ indicates that
group $s$ is retained and $m_{i,s}^{(\ell)}=0$ indicates that it is removed. The mask satisfies
\begin{equation}
\sum\nolimits_{s=1}^{S} m_{i,s}^{(\ell)} = K_i.\nonumber
\end{equation}

We expand this group-level mask to the full value dimension as
\begin{equation}
M_i^{(\ell)}
=
\operatorname{diag}
\left(
m_{i,1}^{(\ell)}I_{d_s},
\ldots,
m_{i,S}^{(\ell)}I_{d_s}
\right)
\in \mathbb{R}^{d_v\times d_v},\nonumber
\end{equation}
where $I_{d_s}$ is the $d_s\times d_s$ identity matrix. The masked
representation is then
\begin{equation}
\widetilde{\mathbf{z}}_i^{(\ell)}
=
M_i^{(\ell)}\mathbf{z}_i^{(\ell)}.\nonumber
\end{equation}
Finally, the routed representation is transformed back to the original value
space:
\begin{equation}
\widetilde{\mathbf{v}}_i^{(\ell)}
=
R^{(\ell)}
\widetilde{\mathbf{z}}_i^{(\ell)}
=
R^{(\ell)}
M_i^{(\ell)}
R^{(\ell)\top}
\mathbf{v}_i^{(\ell)}.\nonumber
%\label{eq:routedv}
\end{equation}
The orthogonal transformation itself is lossless. If all groups are retained,
then $m_{i,s}^{(\ell)}=1$ for every $s$, so $M_i^{(\ell)}=I$ and
$\
\widetilde{\mathbf{v}}_i^{(\ell)}
=
R^{(\ell)}
R^{(\ell)\top}
\mathbf{v}_i^{(\ell)}
=
\mathbf{v}_i^{(\ell)}.
$
Thus, information is removed only through group selection. The groups are not
assigned predefined semantic meanings. They provide a structured space in
which the routing mechanism learns which parts of the value representation to
retain.

\subsection{Learned Latent Role Representation}
\label{sec:roles}

To provide an additional learned signal for value-group selection, RoleSub
associates each token with a low-dimensional latent role representation. For
token $i$ at layer $\ell$, RoleSub computes a vector $\mathbf{a}_i^{(\ell)}$ through a lightweight role estimator as
\begin{equation}
\mathbf{a}_i^{(\ell)}
=
\operatorname{softmax}
\left(
f_{\mathrm{role}}^{(\ell)}
\left(
\mathbf{h}_i^{(\ell)}
\right)
\right)
\in
\mathbb{R}^{C},
\label{eq:latentrole}
\end{equation}
where $C$ is the number of latent role components and
$f_{\mathrm{role}}^{(\ell)}$ is the role-estimation network at layer $\ell$. The vector $\mathbf{a}_i^{(\ell)}$ is a soft representation, so a token may
participate in multiple latent components. These components are not assigned
predefined semantic meanings. Instead, their organization is learned jointly
with the policy through their contribution to value-group routing and action
prediction.

\subsection{Role-Conditioned Value Routing}
\label{sec:router}

For each routed token, RoleSub assigns a routing score to each of the $S$
value groups using three sources of information: the token's current hidden
state, its latent role representation, and the language context.

First, a token-state scoring head maps the hidden state of token $i$ to
group-level routing scores:
\begin{equation}
\boldsymbol{\gamma}_{i,\mathrm{token}}^{(\ell)}
=
f_v^{(\ell)}
\left(
\mathbf{h}_i^{(\ell)}
\right)
\in
\mathbb{R}^{S},\nonumber
\end{equation}
where $f_v^{(\ell)}$ is a learned scoring function. The $s$-th element
$\gamma_{i,\mathrm{token},s}^{(\ell)}$ measures the preference for retaining
value group $s$ based on the current hidden state of token $i$.

The latent role representation provides a second routing signal:
\begin{equation}
\boldsymbol{\gamma}_{i,\mathrm{role}}^{(\ell)}
=
W_{\mathrm{role}}^{(\ell)}
\mathbf{a}_i^{(\ell)}
\in
\mathbb{R}^{S},\nonumber
\end{equation}
where
$W_{\mathrm{role}}^{(\ell)}\in\mathbb{R}^{S\times C}$
is a learned projection from the $C$ latent role components to the $S$
value-group scores. The routing decision also incorporates the current language instruction. Let
$\mathcal{I}_{\mathrm{lang}}^{\mathrm{keep}}$ denote the retained language
tokens. Their hidden states are summarized as
\begin{equation}
\overline{\mathbf{h}}_{\mathrm{lang}}^{(\ell)}
=
\operatorname{Pool}
\left(
\left\{
\mathbf{h}_j^{(\ell)}
:
j\in\mathcal{I}_{\mathrm{lang}}^{\mathrm{keep}}
\right\}
\right),\nonumber
\end{equation}
where $\operatorname{Pool}(\cdot)$ aggregates the language-token hidden states
into a single representation. A learned language scoring function then
produces
\begin{equation}
\boldsymbol{\gamma}_{\mathrm{lang}}^{(\ell)}
=
f_{\mathrm{lang}}^{(\ell)}
\left(
\overline{\mathbf{h}}_{\mathrm{lang}}^{(\ell)}
\right)
\in
\mathbb{R}^{S}.\nonumber
\end{equation}
The three routing signals are combined to produce the routing score
\begin{equation}
\boldsymbol{\gamma}_i^{(\ell)}
=
\boldsymbol{\gamma}_{i,\mathrm{token}}^{(\ell)}
+
\alpha_{\mathrm{role}}
\boldsymbol{\gamma}_{i,\mathrm{role}}^{(\ell)}
+
\alpha_{\mathrm{lang}}
\boldsymbol{\gamma}_{\mathrm{lang}}^{(\ell)},
\label{eq:routingscore}
\end{equation}
where $\alpha_{\mathrm{role}}$ and $\alpha_{\mathrm{lang}}$ control 
the contributions of the latent-role and language-context terms. 

If token $i$ is assigned a budget $K_i$, the router retains the
$K_i$ groups with the highest routing scores. Because the Top-$K_i$ selection is discrete, a straight-through estimator is used during backpropagation. The same routing mechanism is applied to different token types, while their value-group budgets can be controlled separately.

\subsection{Token-Type-Dependent Value Budgets}
\label{sec:budgets}

The routing scores in Equation~\ref{eq:routingscore} determine which value groups are useful for a token. RoleSub additionally allows the number of retained groups to vary across tokens. This is useful because retained tokens
need not require the same amount of value information. Some tokens may carry
information that is more important to the current control decision.

For each token type
$
m
\in
\{
\mathrm{vis},
\mathrm{lang},
\mathrm{prop},
\mathrm{act}
\},
$
let $K_m$ denote the target average number of retained value groups out of
the total $S$ groups. The corresponding total group budget for the retained
tokens of type $m$ in the retained set $\mathcal{I}_m^{\mathrm{keep}}$ is
\begin{equation}
B_m
=
K_m
\left|
\mathcal{I}_m^{\mathrm{keep}}
\right|.\nonumber
\end{equation}
Rather than assigning exactly $K_m$ groups to every token, RoleSub uses the
latent role representation $\mathbf{a}_i^{(\ell)}$ as a token-level priority
signal for distributing $B_m$. We denote this role-driven soft allocation by
\begin{equation}
\left\{
K_i^{(\ell)}
:
i\in\mathcal{I}_m^{\mathrm{keep}}
\right\}
=
\mathcal{B}^{(\ell)}
\left(
\left\{
\mathbf{a}_i^{(\ell)}
:
i\in\mathcal{I}_m^{\mathrm{keep}}
\right\},
B_m
\right),\nonumber
%\label{eq:budgetalloc}
\end{equation}
where $\mathcal{B}^{(\ell)}$ is the budget allocator at layer $\ell$.
The allocator compares the role-derived priorities of tokens within the same
token type and assigns more value-group capacity to higher-priority tokens
and less to lower-priority tokens, subject to
\begin{equation}
0
\leq
K_i^{(\ell)}
\leq
S,
\qquad
\sum\nolimits_{i\in\mathcal{I}_m^{\mathrm{keep}}}
K_i^{(\ell)}
=
B_m.\nonumber
\end{equation}
That is, the average allocation remains $K_m$ even though individual
tokens can have different budgets. When $K_m=S$, all tokens of type $m$ retain their complete value representations and no within-type allocation is necessary. For $K_m<S$, the allocator redistributes the available value capacity among tokens while maintaining $K_m$ for that token type.

\subsection{Training}
\label{sec:training}

We keep the pretrained VLA backbone frozen and jointly train LoRA adapters~\cite{hu2022lora}, the routing functions, and the latent role estimators using the action-prediction objective.

Let $T_a$ denote the number of actions in an action chunk, $\mathbf{y}_t$ the ground-truth action at position $t$, and $\widehat{\mathbf{y}}_t$ the corresponding prediction. The training loss is
\begin{equation}
\mathcal{L}_{\mathrm{act}}
=
\frac{1}{T_a}
\sum\nolimits_{t=1}^{T_a}
\left\|
\widehat{\mathbf{y}}_t
-
\mathbf{y}_t
\right\|_1.
\label{eq:actionloss}
\end{equation}
The  components of the latent role representation $\mathbf{a}_i^{(\ell)}$ are learned jointly with the routing mechanism through their effect on the selected value groups and, ultimately, on the action-prediction loss. %We additionally study optional regularization of the latent representation in Appendix~\ref{sec:ablation_training}.

\section{Experiments}
\label{sec:experiments}

\subsection{Experimental Setup}
\label{sec:exp_setup}

\paragraph{Model.}
We use OpenVLA-OFT-7B~\cite{kim2025oft}, which is built on
OpenVLA~\cite{kim2025openvla}. The underlying OpenVLA architecture combines
a Llama-2-7B language model with fused SigLIP and DINOv2 visual features.
OpenVLA-OFT uses parallel action decoding, continuous action prediction,
action chunking, proprioceptive input, and two camera views. In our
configuration, each control step predicts an eight-action chunk. The multimodal prefix contains 512 visual tokens from two camera views,
together with language, proprioceptive, and action-context tokens. The 
number of language tokens depends on the instruction. %Unless otherwise stated, token-level reduction is applied only to the visual tokens.

\paragraph{Token-level reduction.}
For visual-token reduction, we use the QK-based selection mechanism of the
VLA-ADP implementation~\cite{pei2025vlaadp}. Token importance is computed at
the embedding layer. The retained visual-token budget is divided evenly between the two camera views. We evaluate visual retention ratios
\begin{equation}
r_{\mathrm{vis}}
\in
\{0.015625,\,0.03125,\,0.0625\},\nonumber
\end{equation}
corresponding to 8, 16, and 32 retained visual tokens out of the original 512.

\paragraph{RoleSub configuration.}
The value dimension is $d_v=4096$. For each transformer layer, we use a
fixed orthonormal matrix $R^{(\ell)}$ initialized from a random Stiefel
matrix and kept frozen throughout training. The transformed value
representation is divided into
\(S=16\) groups, 256 dimensions each.

The latent role estimator produces
\(
C=6
\) latent components. The routing score combines the token-state, latent-role,
and language-context terms defined in Sec.~\ref{sec:router}. We set
\begin{equation}
\alpha_{\mathrm{role}}
=
\alpha_{\mathrm{lang}}
=
0.1.\nonumber
\end{equation}
For the main visual-compression experiments, we evaluate
\begin{equation}
K_{\mathrm{vis}}
\in
\{1,2,4\},
\end{equation}
while language, proprioceptive, and action-context values remain
uncompressed. We separately vary $K_{\mathrm{lang}}$ in the language-only
and combined experiments.

\paragraph{Training data.}
We train on the no-op-filtered LIBERO RLDS data~\cite{liu2023libero}.
The Spatial, Object, Goal, and LIBERO-10 suites contain 432, 454, 428, and
379 training episodes, respectively. Image augmentation is enabled during
training.

\paragraph{Optimization.}
We keep the pretrained VLA backbone frozen and jointly train LoRA
adapters~\cite{hu2022lora}, the routing functions, and the latent role
estimators. LoRA is applied to all linear layers with rank 32, scaling 16,
and zero dropout. Training uses AdamW with a learning rate of
$2\times10^{-4}$ and an effective batch size of 8.

The LoRA adapters contain approximately 110.8M trainable parameters out of
7.65B total model parameters. The routing heads add approximately 4.20M
parameters, and the latent role estimator adds approximately 0.55M
parameters. The orthogonal matrices are frozen and therefore introduce no
trainable parameters.

%The main RoleSub results use only the original action-prediction objective in Eq.~\ref{eq:actionloss}. The optional role-prior, sparsity, and alignment losses are evaluated separately in Appendix~\ref{sec:ablation_training}.

\paragraph{Evaluation.}
We evaluate closed-loop task success on the four LIBERO
suites~\cite{liu2023libero}. Each suite contains 10 tasks. Each evaluated
checkpoint is run for 50 trials per task. The simulator configuration and evaluation seed are fixed across compared
methods.

%\paragraph{Checkpoint and replicate aggregation.}
%Aggressive compression can produce considerable variation across training runs. Each completed run is therefore evaluated at multiple late-stage checkpoints. A checkpoint is considered valid only if at least 450 of the planned 500 evaluation episodes complete. For each training run, we retain the best valid late checkpoint. When multiple successful training runs are available for the same configuration, the reported value is the mean of the two strongest runs. Runs with less than 10\% success are treated as failed optimization runs and are excluded from this aggregation. We discuss the limitations of this protocol in Sec.~\ref{sec:limitations}.

\subsection{Baselines}
\label{sec:baselines}

\paragraph{Uncompressed VLA.}
The dense reference is the official OpenVLA-OFT checkpoint for each LIBERO
suite~\cite{kim2025oft}, evaluated using the same configuration as
the compressed models.

\paragraph{Training-free token-only pruning.}
We apply visual-token pruning directly to the finetuned VLA without
additional training. This baseline tests whether the aggressive KV budgets
considered in this work can be reached simply by removing more visual tokens.

\paragraph{Matched-budget trained token-only pruning.}
Because RoleSub is trained jointly with the compressed policy, comparison
against training-free pruning alone would confound the effect of routing with
the effect of adaptation. We therefore construct a trained token-only pruning
using the same LoRA training procedure as RoleSub but without sub-token
routing.

For a RoleSub configuration with visual-token retention ratio
$r_{\mathrm{vis}}$ and mean value-group budget $K_{\mathrm{vis}}$, each
retained visual token uses the full key representation and
$K_{\mathrm{vis}}/S$ of the value representation. The matched token-only
pruning therefore retains
\begin{equation}
r_{\mathrm{matched}}
=
r_{\mathrm{vis}}
\left(S+K_{\mathrm{vis}}\right)/{2S}\nonumber
\end{equation}
of the original visual tokens while keeping their complete key and value
representations. %With $S=16$, this becomes
%\begin{equation}
%r_{\mathrm{matched}} = r_{\mathrm{vis}} \frac{16+K_{\mathrm{vis}}}{32}. 
%\end{equation}
This control isolates whether retaining more token locations with compressed
values provides an advantage over retaining fewer tokens with complete
values at the same visual-KV budget.

\paragraph{Plain sub-token routing.}
We also construct a trained sub-token routing control that retains the same
token and value budgets as RoleSub but removes the VLA-specific routing
structure. It operates directly in the native value basis by setting $R=I$,
removes the latent-role contribution from the routing score, and assigns the
same $K_{\mathrm{vis}}$ groups to every retained visual token. This comparison
isolates the contribution of the orthogonal routing space, latent role
conditioning, and adaptive within-type budget allocation from the general
benefit of reducing values within retained tokens.

\subsection{Visual Compression at Matched KV Budget}
\label{sec:visual_results}

We first evaluate sub-token routing on visual representations. Language,
proprioceptive, and action-context values remain uncompressed. We vary the
number of retained visual tokens and the value-group budget
$K_{\mathrm{vis}}\in\{1,2,4\}$. For each configuration, the trained
token-only pruning uses the same LoRA training procedure and is matched to
the same visual-KV budget. Table~\ref{tab:visual_sweep} reports the results.

\begin{table*}[t]
\centering
\small
\caption{
Visual compression at matched KV budgets. RoleSub retains 8, 16, or 32 of
the original 512 visual tokens and varies the mean retained value-group
budget $K_{\mathrm{vis}}$. The trained token-only pruning retains complete
values but uses fewer visual tokens to match the same visual-KV budget.
}
\label{tab:visual_sweep}
\setlength{\tabcolsep}{5pt}
\begin{tabular}{ccclcccc}
\toprule
Visual keep & $K_{\mathrm{vis}}$ & Visual KV & Method
& Goal & LIBERO-10 & Object & Spatial\\
\midrule
8/512
& 1 & 0.83\% & RoleSub
& \textbf{93.5} & \textbf{74.2} & 95.5 & \textbf{95.6}\\
& & & Token-only
& 85.7 & 73.6 & \textbf{96.5} & 88.3\\

& 2 & 0.88\% & RoleSub
& \textbf{91.8} & \textbf{76.9} & \textbf{95.9} & 96.7\\
& & & Token-only
& 88.7 & 67.7 & 95.2 & \textbf{96.8}\\

& 4 & 0.98\% & RoleSub
& \textbf{93.4} & \textbf{71.8} & 95.8 & \textbf{95.3}\\
& & & Token-only
& 84.7 & 67.7 & \textbf{96.9} & 94.6\\
\midrule

16/512
& 1 & 1.66\% & RoleSub
& \textbf{95.3} & \textbf{83.9} & \textbf{96.5} & \textbf{97.0}\\
& & & Token-only
& 94.6 & 75.3 & 95.4 & 93.7\\

& 2 & 1.76\% & RoleSub
& \textbf{93.4} & \textbf{80.9} & \textbf{95.9} & \textbf{97.7}\\
& & & Token-only
& 91.6 & 72.5 & 94.6 & 96.6\\

& 4 & 1.95\% & RoleSub
& \textbf{93.7} & \textbf{84.5} & \textbf{96.8} & \textbf{96.5}\\
& & & Token-only
& 89.5 & 74.0 & 92.3 & 96.2\\
\midrule

32/512
& 1 & 3.32\% & RoleSub
& \textbf{97.2} & \textbf{88.4} & \textbf{96.4} & \textbf{97.8}\\
& & & Token-only
& 94.8 & 84.0 & 91.7 & 96.8\\

& 2 & 3.52\% & RoleSub
& \textbf{97.5} & \textbf{88.3} & \textbf{97.5} & \textbf{97.8}\\
& & & Token-only
& 96.4 & 84.9 & 97.0 & 97.2\\

& 4 & 3.91\% & RoleSub
& \textbf{96.0} & \textbf{88.0} & \textbf{96.8} & \textbf{97.4}\\
& & & Token-only
& 95.9 & 87.2 & 96.7 & 97.2\\
\bottomrule
\end{tabular}
\end{table*}

Before comparing the two trained methods, we test whether the same aggressive
budgets can be reached by applying token pruning directly to the finetuned
VLA without additional training. At the twelve operating points matched to
the $K_{\mathrm{vis}}=1$ configurations, training-free token pruning produces
zero success in 10 of 12 cases. The only nonzero results are 29.0 on Goal
and 9.2 on Spatial at the largest visual budget. Thus, simply increasing the
amount of token pruning is not viable in the compression rate considered
here. We therefore use the trained token-only model as the primary
matched-budget baseline to compare, so that both methods are adapted under compression.

From the Table~\ref{tab:visual_sweep}, RoleSub outperforms the trained token-only pruning in 33 of 36 matched-budget settings. The largest gains appear on LIBERO-10 and under the smallest visual-token budgets. On LIBERO-10, RoleSub improves over the matched token-only pruning by as much as 10.5 percentage points, while the largest Goal improvement is 8.7 points.

The three exceptions occur at the 8-token setting on the near-saturated
Object or Spatial suites, where the differences are only 0.1--1.1 points.
At the 32-token settings, RoleSub is close to the dense policy across Goal, Object, and Spatial while using only 3.32--3.91\% of the original visual KV. LIBERO-10 remains more sensitive to compression, but RoleSub consistently outperforms the matched token-only pruning across all nine visual configurations.

\subsection{Effect of Role-Conditioned Routing}
\label{sec:role_ablation}

We next compare RoleSub with plain sub-token routing at $K_{\mathrm{vis}}=1$. The plain router uses the same visual-token pruning and value-group reduction budgets, but operates in the native value basis ($R=I$) and removes the latent-role conditioning.  As shown in Table~\ref{tab:role_ablation}, RoleSub has more advantages under the more restrictive visual budgets. With only 8 retained visual tokens, RoleSub improves LIBERO-10 from 59.8\% to 74.2\%, a gain of 14.4 percentage points, and improves Goal from 89.4\% to 95.2\%. At larger visual budgets, the gap narrows as both methods retain more information. Object and Spatial are less discriminative because both methods already operate close to the success ceiling.

\begin{table}[t]
\centering
\small
\caption{
Comparison with plain sub-token routing at $K_{\mathrm{vis}}=1$.
}
\label{tab:role_ablation}
\begin{tabular}{cclcccc}
\toprule
Visual keep & Visual KV & Method
& Goal & LIBERO-10 & Object & Spatial\\
\midrule
8/512 & 0.83\% & RoleSub
& \textbf{93.5} & \textbf{74.2} & \textbf{95.5} & 95.6\\
& & Plain
& 89.4 & 59.8 & 94.6 & \textbf{96.2}\\
\midrule
16/512 & 1.66\% & RoleSub
& \textbf{95.3} & \textbf{83.9} & 96.5 & \textbf{97.0}\\
& & Plain
& 93.2 & 83.4 & \textbf{97.2} & 95.2\\
\midrule
32/512 & 3.32\% & RoleSub
& \textbf{97.2} & \textbf{88.4} & 96.4 & 97.8\\
& & Plain
& 94.8 & 85.8 & \textbf{96.6} & 97.8\\
\bottomrule
\end{tabular}
\end{table}

These results show that plain within-token value compression is not sufficient
to explain the gains of RoleSub. The additional routing structure becomes
most useful when the representation budget is highly constrained, where the
policy is more sensitive to how the limited value capacity is distributed.

\subsection{Language Token and Value Compression}
\label{sec:language}

The above experiments suggest that removing a complete visual token can be much
more destructive than reducing the value capacity of a retained token. We observe an even stronger effect for language. Directly pruning language tokens causes a sharp loss in performance, motivating us to ask whether the language representation can instead be compressed along the value dimension while preserving the complete instruction sequence.

We first evaluate language-token pruning on the Goal suite. The retained tokens keep their complete key and value representations. While mild pruning can be tolerated, but performance degrades rapidly once a larger fraction of the instruction tokens is removed. For example, when 53\% of token are retained, the success rate drop to about only 40\%. These results indicate that the policy depends strongly on preserving the language-token sequence.

We therefore test a different compression axis. Instead of removing language
tokens, we retain every language token and reduce only its value width using
$K_{\mathrm{lang}}\in\{8,4,2,1\}$. Visual, proprioceptive, and action-context representations remain uncompressed. Since keys remain full, the language-KV fraction for a value-group budget $K_{\mathrm{lang}}$ is
\begin{equation}
\rho_{\mathrm{lang}}
=
\left(1+K_{\mathrm{lang}}/S\right)/{2}.
\end{equation}
As shown in Table~\ref{tab:language}, in contrast to token pruning, language value compression is essentially lossless across the entire sweep. Even at $K_{\mathrm{lang}}=1$, where each language token retains only one of 16 value groups, performance remains at the dense-policy level on all four suites. These results indicate that language-token positions are critical to the VLA policy, whereas much of the value width associated with those tokens is redundant.

\begin{table}[t]
\centering
\small
\caption{
Language-only value routing with $S=16$ value groups. 
}
\label{tab:language}
\begin{tabular}{ccccccc}
\toprule
$K_{\mathrm{lang}}$
& Language V
& Language KV
& Goal
& LIBERO-10
& Object
& Spatial\\
\midrule
8 & 50.0\%  & 75.0\%  & 97.0 & 94.8 & 98.2 & 98.2\\
4 & 25.0\%  & 62.5\%  & 98.0 & 94.6 & 98.2 & 98.4\\
2 & 12.5\%  & 56.25\% & 98.4 & 93.8 & 98.4 & 98.4\\
1 & 6.25\%  & 53.13\% & 98.0 & 94.8 & 98.4 & 98.0\\
\midrule
Dense & 100\% & 100\% & 98.8 & 94.8 & 97.8 & 98.2\\
\bottomrule
\end{tabular}
\end{table}

\subsection{Combined Visual and Language Compression}
\label{sec:combined}

We next combine visual and language compression in the same model. Following
the visual experiments, we retain 8, 16, or 32 of the 512 visual tokens and
apply sub-token routing to the retained visual tokens with
$K_{\mathrm{vis}}=1$. All language tokens are retained, while their value
representations are compressed with $K_{\mathrm{lang}}=1$. Proprioceptive
and action-context representations remain uncompressed. The resulting total
KV fractions are 9.2\%, 9.9\%, and 11.3\%, respectively.

\begin{table*}[t]
\centering
\small
\caption{
Combined visual and language compression with
$K_{\mathrm{vis}}=K_{\mathrm{lang}}=1$.
``Visual-only'' uses the same visual-token retention and
$K_{\mathrm{vis}}=1$, but keeps the language values uncompressed.
}
\label{tab:combined}
\begin{tabular}{lcccccc}
\toprule
Suite & Visual keep
& Combined KV & Combined
& Visual-only KV & Visual-only & $\Delta$\\
\midrule
Goal
& 8/512  & 9.2\%  & 92.7 & 15.4\% & 93.5 & $-0.8$\\
Goal
& 16/512 & 9.9\%  & 93.6 & 16.1\% & 95.3 & $-1.7$\\
Goal
& 32/512 & 11.3\% & 94.3 & 17.5\% & 97.2 & $-2.9$\\
\midrule
LIBERO-10
& 8/512  & 9.2\%  & 64.2 & 15.4\% & 74.2 & $-10.0$\\
LIBERO-10
& 16/512 & 9.9\%  & 72.7 & 16.1\% & 83.9 & $-11.2$\\
LIBERO-10
& 32/512 & 11.3\% & 77.6 & 17.5\% & 88.4 & $-10.8$\\
\midrule
Object
& 8/512  & 9.2\%  & 96.2 & 15.4\% & 95.5 & $+0.7$\\
Object
& 16/512 & 9.9\%  & 95.6 & 16.1\% & 96.5 & $-0.9$\\
Object
& 32/512 & 11.3\% & 95.9 & 17.5\% & 96.4 & $-0.5$\\
\midrule
Spatial
& 8/512  & 9.2\%  & 96.3 & 15.4\% & 95.6 & $+0.7$\\
Spatial
& 16/512 & 9.9\%  & 96.6 & 16.1\% & 97.0 & $-0.4$\\
Spatial
& 32/512 & 11.3\% & 96.4 & 17.5\% & 97.8 & $-1.4$\\
\bottomrule
\end{tabular}
\end{table*}

As shown in Table~\ref{tab:combined}, Object and Spatial tasks remain very close to
their visual-only results after language compression is added, with differences between $-1.4$ and $+0.7$ percentage points. Goal also remains relatively stable, with losses of 0.8--2.9 points. That is, visual and language compression can be reasonably combined on these tasks to reduce the total KV to about 9--11\%.

LIBERO-10 is quite sensitive to combining the two forms of compression.
Adding $K_{\mathrm{lang}}=1$ language-value compression reduces success by
10.0--11.2 points relative to the corresponding visual-only settings. This
differs from the language-only experiment in Table~\ref{tab:language}, where
$K_{\mathrm{lang}}=1$ causes no measurable loss when the visual representation is not compressed. The result shows that the effect of language compression depends on how much visual information is also retained. We examine this interaction further by varying the language value budget in the next subsection.

\subsection{Language Budget on LIBERO-10}
\label{sec:long}

We vary the language value budget while keeping the visual routing configuration fixed at $K_{\mathrm{vis}}=1$. We evaluate $K_{\mathrm{lang}}\in\{1,8,16\}$ for each of the 8-, 16-, and 32-token visual settings. Here, $K_{\mathrm{lang}}=16$ retains the full language value representation.

\begin{table}[t]
\centering
\small
\caption{
LIBERO-10 success rate with different language value budgets. $K_{\mathrm{vis}}=1$ in all settings.
}
\label{tab:langsweep}
\begin{tabular}{lccc}
\toprule
Language budget & 8 visual tokens & 16 visual tokens & 32 visual tokens\\
\midrule
$K_{\mathrm{lang}}=1$
& 64.2 & 72.7 & 77.6\\
$K_{\mathrm{lang}}=8$
& 68.4 & 79.7 & 84.8\\
$K_{\mathrm{lang}}=16$
& 83.6 & 86.8 & 87.6\\
\bottomrule
\end{tabular}
\end{table}

As shown in Table~\ref{tab:langsweep}, increasing the language value budget improves LIBERO-10 performance at all three visual-token settings. Moving from $K_{\mathrm{lang}}=1$ to $K_{\mathrm{lang}}=8$ improves success by 4.2, 7.0, and 7.2 percentage points for 8, 16, and 32 visual tokens, respectively. With 32 visual tokens, $K_{\mathrm{lang}}=8$ reaches 84.8\%, compared with 87.6\% when the full
language values are retained.

The results also show an interaction between the visual and language budgets. When only eight visual tokens are retained, increasing the language budget from 8 to 16 groups gives a much larger improvement, from 68.4\% to 83.6\%. In contrast, with 32 visual tokens, eight language groups already recover most of the performance obtained with full language values. This suggests that the visual and language representations are not used independently by the policy. When information from one modality is strongly reduced, the policy appears to rely more heavily on the other. The observed coupling between the two representations is not explicitly modeled in RoleSub and could be explored in future work, for example through joint cross-modal
budget allocation or routing.

\section{Limitations}
\label{sec:limitations}

RoleSub learns a latent role representation for routing, but the learned
components are not guaranteed to correspond to distinct or interpretable
semantic roles. The current experiments evaluate their usefulness for compression rather than their semantic meaning.

The visual and language experiments also reveal an interaction between the two modalities. In particular, aggressive language compression is nearly lossless when applied alone but becomes more costly on LIBERO-10 when the visual representation is also heavily compressed. RoleSub currently assigns visual and language budgets separately rather than explicitly optimizing their joint allocation. Modeling this cross-modal dependence is an important direction for future work.

Finally, this work evaluates reduction in retained KV rather than end-to-end inference latency. Realizing the corresponding runtime benefit requires efficient implementations of sub-token routing and sparse value computation.

\section{Conclusion}
\label{sec:conclusion}

We presented RoleSub, a sub-token routing method for compressing the KV representations of VLA policies. Rather than relying only on token removal, RoleSub retains more token locations while reducing the value representation within each retained token. A learned latent role representation, an orthogonal value-space decomposition, and token-type-dependent budgets guide how the available value capacity is allocated.

On OpenVLA-OFT-7B across the four LIBERO suites, RoleSub consistently
outperforms trained token-only pruning at matched visual-KV budgets, especially
under aggressive compression. The experiments also show a clear difference
between token and value compression: language tokens are difficult to remove,
while their value representations can be compressed much more aggressively.

By combining visual and language compression, RoleSub reduces the total KV to
9.2--11.3\% of the original while preserving strong performance on most tasks. The results further indicate that visual and language compression are coupled, particularly for long-horizon control. Overall, RoleSub shows that compressing representations within retained tokens provides an effective complement to token-level reduction for efficient VLA policies.

\bibliographystyle{plainnat}
\bibliography{ref}

\end{document}